\documentclass[twocolumn]{article}
\usepackage[top=2cm, bottom=2.5cm, left=2cm, right=2cm]{geometry}
\usepackage{times,textcomp}
\usepackage[T1]{fontenc}
\usepackage[utf8]{inputenc}
\usepackage{amssymb,amsmath}
\usepackage{enumitem}
\usepackage{eurosym}

\usepackage{titlesec}

\usepackage{titling}
\newcommand{\subtitle}[1]{
	\posttitle{
		\par\end{center}
	\begin{center}\large#1\end{center}
	\vskip0.5em}
}

\usepackage{titlesec} 
\titleformat{\section}{\large\bfseries}{\thesection}{1em}{\MakeUppercase} 
\titleformat{\subsection}{\large\bfseries}{\thesubsection}{1em}{}
\titlespacing*{\section}{0pt}{\parskip}{0pt}
\titlespacing*{\subsection}{0pt}{\parskip}{0pt}

\usepackage{longtable,booktabs}
\usepackage{graphicx,grffile}
\usepackage{fancyhdr}

\title{\huge\bfseries Using NLP on news headlines to predict index trends}

\usepackage{authblk}

\author{Marc Velay}
\author{Fabrice Daniel}

\affil{\small Artificial Intelligence Department of Lusis, Paris, France\\fabrice.daniel@lusis.fr\\http://www.lusis.fr}

\date{June 2018}

\begin{document}

\maketitle

\begin{abstract}
This paper attempts to provide a state of the art in trend prediction using news headlines. We present the research done on predicting DJIA\footnote{Dow Jones Industrial Average} trends using Natural Language Processing. We will explain the different algorithms we have used as well as the various embedding techniques attempted. We rely on statistical and deep learning models in order to extract information from the corpuses.
\end{abstract}


\section{Introduction}

Traders generally look up information about the companies they are looking to buy shares into, for long and short trading. A frequent source of information are news medias, which provide updates about the company's activities, such as expansion, better/worse revenues than expected, new products and much more. Depending on the news, traders can determine a bearish/bullish trend and decide to invest in it.  

We may be able to correlate overall public sentiment towards as a company and its stock price: Apple is generally well liked by the public, receives frequent news coverage of its new products and financial stability and its stock has been growing steadily. These facts may be correlated but the first may not cause the second, we will analyze if news coverage can be used to predict market trend. To do so, we will analyze the top 25 news headline of each open-market day from 2008 to late 2015 and try to predict the end-of-day value of the DJIA index for the same day. The theory behind predicting same day values is that traders will respond to news quickly and thus the market will adjust within hours of release. Therefore in a single business day, if the news is spread during business hours, its effects may be measured before the closing bell of the market. 

The motivation behind this analysis is that humans take decisions using most of the available information. This usually takes several minutes in order to discover new information and take a decision. An algorithm is capable of processing gigabytes of text from multi-source streams in seconds. We could potentially exploit this difference in order to create a trading strategy.

NLP\footnote{Natural Language Processing} techniques can be used to extract different information from the headlines such as sentiments, subjectivity, context and named entities.
We extract indicator vectors using each of these techniques, which allow us to train different algorithms to predict the trend. To predict these values, we can use several techniques which should be well suited for this type of information: Linear regression, Support Vector Machine, Long Short-Term Memory recurrent neural network and a dense feed-forward (MLP) neural network. We included the techniques used by Bollen et al (2010) \cite{Bollen}, which resulted in state-of-the-art results. We will also analyze the techniques used in other studies with a similar context \cite{twitter-sent} \cite{breaking-news}.

\section{Information in headlines}

Latent Sentiment Analysis is done by building up a corpus of labelled words which usually connote a degree of positive or negative sentiment. We can extend the corpus to include emoticons (i.e. ``:-)'' ) and expressions, which often correlate to strong emotions. Naive sentiment analysis consists of a lookup of each word in the sentence to be analyzed and the evaluation of a score for the sentence overall. This approach is limited by its known vocabulary, which can be mitigated by context analysis and the introduction of synonyms. The second limitation is sarcasm, which is prevalent in twitter feed analysis. The sentiment inferred by the words is opposed to the sentiment inferred by the user. This is mitigated by techniques detecting sarcasm which lead to a polarity flip of such tweets. 

Sentiment analysis gives insight on how favorable the media is and maybe the bias traders may have towards buying or selling.

Another NLP technique which gave promising results was context analysis. This is a recent deep learning approach where you rely on a large corpus of text in order to learn and predict the words around a target. You can then deduce in what context it usually appears. The result is a vector representing each word. Other vectors with little distance are usually synonyms. The representation also allows us to do algebra, such as the famous ``king - man + woman = queen''

Learning this representation offers the possibility of associating a specific context with a bullish or bearish market. 
\section{Possible approaches}

To accomplish our goal, we need to combine several techniques for preprocessing, word vectorization and prediction. 

To preprocess the text, we can remove stop words. Those are very common word that are required for correct grammar but add little to sentiment or context analysis. The list includes ``the'', ``a'' and `` and''. Another filter would be for named entities (NER\footnote{Named Entity Recognition}), such as ``IMF'', ``Trump'' or ``France''. By selecting organizations, people and countries we can either remove them for sentiment analysis, as they would otherwise be counted as neutral and ``smooth'' out the effect of other words. They could be used in context analysis as having a very characteristic importance, such as a federal reserve announcing rates for the coming year having a very specific impact on the S\&P 500 index. The problem is that this importance needs to be extracted by a human with general knowledge of the topic. This solution removes the possibility for a fully autonomous system and adds a layer of uncertainty due to human error. We also remove all punctuation from the corpus as our current model does not take them into account. Analyzing the headlines with punctuation included might yield results, but we have not found any literature which pointed towards that result, as of writing.

There exists several word embedding techniques such as Word2Vec, TF-IDF, Bag-of-words \cite{Embedding} and those can be combined with N-grams in order to augment the information they contain \cite{augmentation}. We have tested those three techniques, with and without using N-grams. Hereafter, we will explain the concepts behind those techniques.

After having cleaned the text by removing stop words and named entities, we need to convert the remaining words to vectors of fixed length to compose the dataset used by the machine learning algorithms. The texts are often of varying length, thus we can not simply encode each word and then concatenate them, without losing information. In the case of Word2Vec, which builds a vector representing of the context for each word, we can do a sum of vector, which is the best solution compared to mean and sum of distances, by a margin of 4-5\%.

TF-IDF\footnote{Term Frequency-Inverse Document Frequency} is the number of times a word appears in the top 25 headlines of the day, divided by the amount of times it appeared in the corpus. We then have a vector the size of our vocabulary, with values between 0 and 1.

Bag-of-words simply counts the number of occurrences of a word in the headlines. The algorithm produces a vector of the size of our vocabulary. The values are strictly positive integers.

An N-gram tries to find the most frequent groups of n words. The result is a n amongst m vector, where n is the size of the groups and m is the size of the vocabulary. This could be combined with the Word2Vec technique.

The news headlines are represented as vectors, used to predict an upward or downward trend according to the features present in the vector. The output depends on the algorithm selected, either a one-hot vector for each class or a boolean output. The following algorithms can be considered for the prediction.   

Logistic regression are potential algorithms when used with an activation function that outputs either of the binary target classes.  

We used an SVM\footnote{Support Vector Machine} due to its ability to learn large multi-dimensional vectors. 

We attempted the use of a Gaussian Naive Bayes classifier, popular as a basic text classification technique.

We attempted the use of a LSTM\footnote{Long Short-Term Memory}, but the best results we obtained were made by the model overfitting the training data and not learning a relation between the input and output. We did not improve on randomness using an LSTM.

The last algorithm we tried was an MLP\footnote{Multi-Layer Perceptron}, which did manage to extract a link between the news and the index's value.

In the end, we used almost all the combinations of embedding and prediction algorithms, in order to find the best results.
\section{Results}

After testing most of the combinations of embedding and algorithms, we found that results rarely improved on luck. Deep learning algorithms especially had difficulty figuring out the link between the headlines and the index trend. These predicted single values for whatever input was provided, generally around 0.5, which shows very low confidence in the selected class. The best results we found were using linear regression on bag-of-words vectorized data, which had a 57\% accuracy. The results using this configuration can be found in Table 1.  We believe this result is due to the algorithms not properly extracting information from the large vectors. The headlines are also very context dependent, with recurrent actors for periods of times, before becoming irrelevant. This means that our models should be trained on short time frames, unless it learns information that will not help predict future trends. This limitation also implies that the system is unable to react to breaking news due to not knowing enough about its context to correctly predict a trend. 

\begin{table}[ht]
	\centering
	\begin{tabular}{lr}
		\hline
		{\textbf{Algorithm}}         & {\textbf{Accuracy}} \\ \hline
		\textbf{Logistic Regression} &       \textbf{0.57} \\
		Linear Discriminant Analysis &                0.51 \\
		K-Nearest Neighbors          &                0.46 \\
		Decision Tree Classifier     &                0.49 \\
		Support Vector Machine       &                0.53 \\
		Random Forrest               &                0.50 \\
		Extreme Gradient Boosting    &                0.52 \\
		Naive Bayes                  &                0.53 \\
		LSTM                         &                0.55 \\
		MLP                          &                0.53 \\ \hline
	\end{tabular}
	\caption{Per algorithm validation accuracy of trend prediction using news headlines} 
	\label{tab:results}
\end{table}

We found that multimodal \cite{multimodal} \cite{sarcasm} learning does not provide much greater results when using sentiment and subjectivity indicators. Financial indicators also provided a very slight boost, but the three combined indicators did not manage to be marginally better than luck.

We could improve the solutions by providing human-featured data, such as named entity recognition linked to a more or less important effect. We could also provide pointers as to what sort of news is good or bad, but these additions would imply a semi-autonomous system that could barely match the human that created them. 
\section{Problems encountered}

The core concepts behind the data that we use is biased. We use world news from various outlets. Their main goal is to sell papers, and thus their reporting is oriented in that way. Humans in general have a negativity bias and papers aim to use this to their advantage by publishing negative stories, such as a war breaking in the middle east, a fact reported every few years between 2008 and 2015. Most headlines are objective in phrasing, so there are few samples where subjectivity is not null. The few headlines that are not objective mostly contain negative sentiments, which skews the data towards a negative sentiment.

When we analyze the context of a headline, it is very often relevant to a specific period in time. There a recurrent occurrences of people during a few year before they disappear. The model can therefore not be relied upon for a long extent of time, we should train on 9 months and test on a maximum of 3 months, for information learnt during training to still be relevant for the predictions. Less than 2 months is also of little use due to the low amount of data points tested.

There is a statistical problem linked to the data we chose: if you pick another era, there would most likely be very different results. This is due to several factors, including the rise of broadband internet, which allows users to get more news, faster, as well as creating algorithms very much like this one which sift through gigabytes of data in order to make decisions. As these algorithms become more popular, strategies will be created to counter them, and the correlation level will go down.

Most resources used in this study are vocabulary dependent and rely on humans to update their corpus to take into account new words, expressions and relevant individuals. This implies that the machine learning models on which they rely need to be updated frequently. If the root of our algorithm changes, we need to retrain all our models from scratch again, as on-line learning would not work due to vectors of varying lengths.
\section{Conclusion}

In this study we analyzed several techniques used in predicting market trends using news articles. We detailed different encoding techniques using word embedding. We suggested the use of classification algorithms in order to process the news articles and predict the future trends of the DJIA index.

Our study resulted in the finding that, over a long period of time, there was little to no correlation between news sentiment and the DJIA index trends. When we looked at different word embedding techniques to predict the trend, we found that the results were only slightly better than random. 

The following suggestions might improve the results obtained, but there are still flaws in the idea that a model trained on past news could predict future trends. These flaws can not be resolved until future Machine Learning models are able to learn broad contexts of un-related events, which humans rely on the process new information, in particular news articles. 

Using archives of tweets, we could train a model on the sentiment towards a specific company which depends on popular opinion as a business, such as airlines and Tesla. This strong apriori might prove efficient to predict trends, compared to using headlines to predict a national index. In theory, there should be a stronger correlation.

We should change the training and testing timeframes compared to the current 6 years of training for 1 year of testing distribution. Most of what is learnt will not be used again. We could instead do a moving window of 9+3 months and test our hypothesis on that reduced scale, where market behavior is more likely to be similar.

\end{document}